\documentclass[journal]{IEEEtran}
\usepackage{array}
\usepackage{textcomp}
\usepackage{stfloats}
\usepackage{url}
\usepackage{verbatim}
\usepackage{cite}
\usepackage{amsmath,amssymb,amsfonts}
\usepackage{graphicx}
\usepackage{textcomp}
\usepackage{float}
\usepackage[noend]{algpseudocode}
\usepackage{color}
\usepackage{algorithm}
\usepackage{algorithmicx}
\usepackage{booktabs}
\usepackage{threeparttable}
\usepackage{multirow}
\usepackage{multicol}
\usepackage{subfigure}
\usepackage[table,xcdraw]{xcolor}
\usepackage{epstopdf}

\begin{document}

\title{\huge BG-GAN: Generative AI Enable Representing Brain Structure-Function Connections for Alzheimer's Disease}

\author{Tong Zhou, Chen Ding, Changhong Jing, Feng Liu, Kevin Hung, Hieu Pham, Mufti Mahmud, Zhihan Lyu, Sibo Qiao, Shuqiang Wang, Kim-Fung Tsang, \textit{Fellow, IEEE}
	
	\thanks{Tong Zhou and Chen Ding contributed equally to this work}
    \thanks{Corresponding author:Shuqiang Wang, sq.wang@siat.ac.cn}
	\thanks{Tong Zhou, Chen Ding, Changhong Jing, Shuqiang Wang and Kim-Fung Tsang are with the Shenzhen Institutes of Advanced Technology, Chinese Academy of Sciences, Shenzhen, 518055, China; Tong Zhou and Shuqiang Wang are also with University of Chinese Academy of Sciences, Beijing 100864, China}
    \thanks{Feng Liu is with the Department of Systems and Enterprises, Stevens Institute of Technology, USA}
    \thanks{Kevin Hung is with School of Science and Technology, Hong Kong Metropolitan University, Hong Kong}
    \thanks{Hieu Pham is with College of Engineering and Computer Science, and VinUni-Illinois Smart Health Center, VinUniversity, Vietnam}
    \thanks{Mufti Mahmud is with Department of Information and Computer Science, SDAIA-KFUPM Joint Research Center for AI, Interdisciplinary Research Center for Biosystems and Machines, King Fahd University of Petroleum and Minerals, Dhahran, Saudi Arabia}
	\thanks{Zhihan Lyu is with Department of Game Design, Faculty of Arts, Uppsala University, Sweden}
    \thanks{Sibo Qiao is with the School of Software, Tiangong University, Tianjin 300387, China}
	
}

\maketitle

\begin{abstract}
The relationship between brain structure and function is critical for revealing the pathogenesis of brain disorders, including Alzheimer's disease (AD). However, mapping brain structure to function connections is a very challenging task.
In this work, a bidirectional graph generative adversarial network (BG-GAN) is proposed to represent brain structure-function connections.
Specifically, by designing a module incorporating inner graph convolution network (InnerGCN), the generators of BG-GAN can employ features of direct and indirect brain regions to learn the mapping function between the structural domain and the functional domain.
Besides, a new module named Balancer is designed to counterpoise the optimization between generators and discriminators.
By introducing the Balancer into BG-GAN, both the structural generator and functional generator can not only alleviate the issue of mode collapse but also learn complementarity of structural and functional features.
Experimental results using Alzheimer's Disease Neuroimaging Initiative (ADNI) dataset show that both generated structure and function connections can improve the
identification accuracy of AD. The experimental findings suggest that the relationship between brain structure and function is not a complete one-to-one correspondence. They also suggest that brain structure is the basis of brain function, and the strong structural connections are majorly accompanied by strong functional connections.

\end{abstract}


\section{Introduction}
Alzheimer's Disease (AD) has placed a huge burden on patients' families and society as a whole. Symptoms of AD mainly include memory impairment, aphasia, apraxia, cognitive impairment, impairment of visuospatial skills, executive dysfunction, and personality and behavior changes. Although most people experience some degree of memory loss in old age, memory loss is much more severe in people with AD than in normal individuals. The term Mild Cognitive Impairment (MCI) was created to differentiate them. Some MCI patients can remain stable for years, while some subjects with AD-like brains change inevitably progress to AD. The prevalence of AD is increasing rapidly due to the fact that the proportion of the population aged 65 and over is growing faster than any other age group worldwide in recent years.
Current drugs can relieve symptoms but do not prevent disease progression. Learning the relationship between brain structure and function can help doctors differentiate between AD patients and normal people (NC). Moreover, it can also help researchers to figure out the cause of the worsening condition\cite{goedert2006century,mattson2004pathways,zou2024long,mucke2009alzheimer}.

Recent advancements in artificial intelligence (AI)\cite{dong2022crowd,ayata2018emotion,sundaravadivel2018smart,xu2024neural} have significantly enhanced the capabilities of neuroimaging analysis, making it more reliable and sensitive than traditional cognitive assessments in detecting AD. Images of different modalities, such as structural Magnetic Resonance Imaging (sMRI), functional Magnetic Resonance Imaging (fMRI), diffusion tensor imaging (DTI) and so on, are well suited to assist physicians in diagnosing patients with the disease \cite{tong2017multi,zhang2012alzheimer,wang2018automatic,wang2017automatic}. Moreover, with the advantage of high temporal and spatial resolution, each of these tools can identify differences between AD patients and NC well.
In the realm of AI-driven diagnostic approaches\cite{cong2022boundary,qiu2022joint,maji2021ikardo,lim2012efficient}, individual modality is used to study diagnosis between AD and MCI. Some used sMRI brain neuroimages to classify AD and NC \cite{lei2020deep,wang2019ensemble,wang2020ensemble,you2022fine,wang2020diabetic,yu2022morphological}. Some used fMRI data \cite{wang2006changes} and some used DTI data \cite{nir2013effectiveness}. Due to the limitations of computing power and algorithm at that time, they couldn't utilize the complementary information between multiple modalities' data\cite{shaffi2024comriad}. The complex relationships between brain structure and function could not be discovered clearly. The results of diagnosis are not so good as the current methods using multiple modalities images. Now, with the proliferation of artificial intelligence technologies, it is possible to integrate data from multiple imaging modalities.

\section{Related Works}
In recent years, more and more researchers have begun to use images of multiple modalities to explore the potential relationship between structure and function\cite{zong2024new}, especially how the emergence and disappearance of functional and structural connections interact with each other. To tackle the above difficulties, many models have been proposed to study AD-related knowledge with data of multiple modalities.
Some models\cite{pan2024decgan} have been developed to learn the relationship between structure and function to explore the reason why a person's brain structure is fixed in a certain period of time, but it can perform different functions.

Zhang et al. \cite{zhang2020recovering} recovered the structural connectivity via Multi-GCN from functional connectivity. The recovered result that functional information can predict structural information, revealed that functional information contains all the structural information.
Wang et al. \cite{wang2020understanding} proposed an algorithm called CDA and the empirical structural connections (SC) can be reliably predicted based on the direct anatomical relationships, indirect pathways, and module topology interacting with one another forming temporally dependent functional connections (FC).
Based on the fact that related brain regions with strong functional connectivity can exist without structural connectivity, but the strong functional connectivity can be observed in the brain regions with the strong structural connectivity, there is a many-to-one functional-structural mode.

However, not all researchers agree with the above views.
Abdelnour et al.\cite{abdelnour2018functional} tried to use mathematical methods to derive a relationship between SC from DTI and FC from fMRI and found that using the approach via (structural) Laplacian spectral, FC and SC shared eigenvectors and their eigenvalues are exponentially related.
Honey et al.\cite{honey2009predicting} believed that although resting state functional connectivity is variable, its strength, persistence and spatial statistics are constrained by the large-scale anatomical structure.
Messé et al.\cite{messe2020parcellation} explored the spatial consistency of the relationship between the structure and function of the human brain. The result showed that brain regions can affect the relationship between structural and functional connections. Different division methods will affect the relationship between structure and function.
Simon et al.\cite{leuchter1997brain} discussed the influence of different brain regions on structure and function and proved its effectiveness.
Koch et al. \cite{koch2002investigation} examined the correlation of spontaneous fluctuations of brain voxel BOLD signals in the cortical gyrus of the same hemisphere and found that there is a positive correlation between function and structure.
Annen et al. \cite{annen2016function} explored the relationship of function fluorodeoxyglucose FDG-PET metabolism and structure MRI-diffusion-weighted images (DWI). The fact that the patients' local metabolism and white matter integrity were significantly reduced, showed that a stronger relationship of function-structure exists in most areas of Default mode networ.
Straathof et al. \cite{straathof2019systematic} combined multiple articles and draw a conclusion: the structural and functional connection strength of the mammalian brain is positively correlated with the connection strength of the diffuse structure on the macroscopic scale, and positively correlated with the connection strength of the neuron trace structure on the mesoscale.

Based on above analysis, the following simple assumption can be summarized: there is indeed a relationship between structure and function. However, very few studies focus on the relationship between structure and function by employing complementary features between brain structure and function.

\section{Contributions}
In this paper, a novel end-to-end framework named BGGAN is proposed where latent vectors and graph information are effectively utilized and jointly optimized in both brain structural domain and functional domain. A hypothesis that there is some potential relationship between brain structure and function is proposed.

The main contributions of this paper can be summarized as follows:

\begin{itemize}
\item A novel framework called BGGAN is proposed by introducing the bidirectional mapping mechanism. Features of brain structure and brain function can be extracted mutually. Both the structural generator and functional generator can learn the complementary features between brain structure and brain function, which play a vital role in analyzing the relationship of brain structure-function.

\item A new form of graph convolution named InnerGCN is designed. By extending graph convolution, the developed InnerGCN can deal with the third-order tensor while the traditional graph convolution can only process the brain connectivity matrix of a certain modality, namely second-order tensor. By extracting the features of different modalities using circle convolution mechanism, the InnerGCN can learn the
    brain structure-function features which are more close to the distribution of source domain.

\item A new module named Balancer is proposed. By introducing the Balancer into BGGAN, the proposed model can not only alleviate the issue of mode collapse but also efficiently learn complementarity of structural and functional features. The generated connections with the Balancer are more detailed and stable than those without the Balancer.
\end{itemize}

The rest of this paper is organized as follows. In section \ref{Sec:Method}, BGGAN is reviewed and the corresponding details for constructing the model are presented. In section \ref{Sec:Experiments}, the ADNI dataset is used to train the BGGAN model and perform bidirectional experiments between structure and function. The generated data are analyzed to reflect the changes in brain structure and function.

\begin{figure*}[htb]
	\centering
	\includegraphics[width=1.0\linewidth]{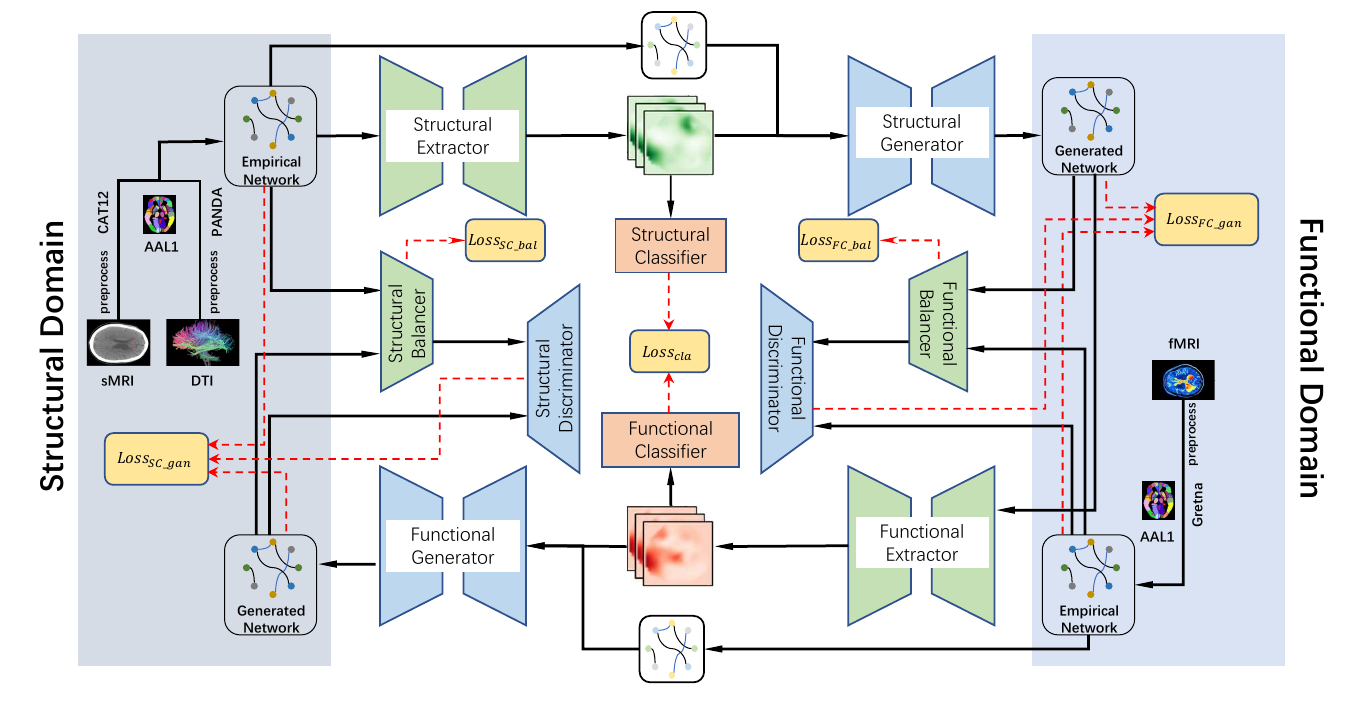}
	\caption{The proposed BGGAN involves two domain: structural domain and functional domain. BGGAN contains two extractors, two generators, two discriminators and two Balancer modules. The training steps are as following:
	(1) Firstly, extract the brain structural features based on sMRI and DTI data with structural extractor, and extract the functional features based on fMRI data with functional feature's extractor.
	(2) Secondly, input the structural and functional features to structural and functional classifiers. The classification results are used to reflect the ability of the feature extractor and guide the learning of the extractors.
	(3) Thirdly, the structural and functional generators use the structural and functional feature matrices from the first step to perform bidirectional mapping between structural and functional domain.
	(4) Fourthly, the module Balancer merges the structural and functional connection matrices to reduce the gap between the source data and the target data.
	(5) Fifthly, structural and functional discriminators evaluate the differences between the generated data and the source data. The above steps describe the process from structure to function, and then reverse the generated function back to the structure. In fact, there is also the process of generating from function to structure, and then from structure to function, which is the same as before.}
	\label{fig:Fig1}
\end{figure*}

\section{Method}
\label{Sec:Method}
\subsection{Overview}
As shown in Fig.\ref{fig:Fig1}, the proposed framework consists of three modules: the Generator modules, the Discriminator modules and the Balancer modules.
 Generators contain feature extractors and brain connection generators. Feature extractors try to learn ninety brain regions' features with multiple modalities. InnerGCN layers first utilize multiple modalities' information more comprehensively. And then, the fully connected layers and the soft-max layers are appended behind the feature extractors to combine the multiple modalities' features for classification. Effective classification results can demonstrate that the InnerGCN layers can map the characteristics of different categories of subjects to corresponding areas in the latent space. The brain connection generators are then able to perform a bidirectional mapping between brain structure and function based on the features in the latent space.
 Discriminators are used to detect the authenticity between the generated data and the source data. The output of the discriminators is used to guide the generators to make the generated data distribution close to the source data distribution.
 The Balancer modules try to reduce the performance gap between the generators and the discriminators. The source data of the discriminator is doped with the information of empirical data from another domain through the Balancer module. Based on this, the gap between the generated data and the source data can be reduced, and the generators can learn the details of the source data distribution more smoothly.

 The BGGAN can not only increase the model’s generalization effect, but also alleviate the modal crash issue. Unlike the traditional GAN model, BGGAN can generate the structural and functional connections at the same time. The Balancer designed in the BGGAN can assist them to balance the performance of the generators and discriminators in a more reasonable growth. The details of BGGAN are told in Section \ref{Sec:section2.B} .

\subsection{BGGAN for mapping domain}
\label{Sec:section2.B}
 Generative adversarial networks \cite{yu2021tensorizing,hu2020medical,yu2020multi,zuo2021multimodal,pan2021characterization} commonly consist of two modules: the module generator and the module discriminator. The generator tries to map latent features $z \sim p(z)$ from the noise or from the prior distribution to the distribution of the real domain. The discriminator aims to distinguish between the real data and the fake data. They are trained by the following min-max optimization Eq. \ref{eq:eq1}:
 \begin{equation}
	\begin{split}
		\min \limits_G \max \limits_D V(D, G) = \mathbb{E}_{x\sim p_{data}(x)}[logD(x)] +\\ \mathbb{E}_{z \sim p_z(z)}[log(1-D(G(z)))].
	\end{split}
	\label{eq:eq1}
 \end{equation}

 Compared with traditional GAN framework using image convolution form, BGGAN in this paper uses the graph convolution as the calculation method. Just as shown in Fig.\ref{fig:Fig1}, the input of generators is the brain connection matrices of multiple modalities, denoted as $\mathcal{G}=\{V, E\}$. $V \in \mathcal{R}^{N \times D \times R}$ and $E \in \mathcal{R}^{N \times N \times R}$. Here, $N$ is the number of brain regions, $D$ is the dimension of feature matrix and $R$ is the number of modalities.
 However, the traditional GCN method can not make full use of the information of multiple modal data. The input of traditional GCN can only be a matrix. When the input is multiple modalities data, parts of the information will be discarded. The traditional GCN method doesn't take into account the relationship of different modalities. InnerGCN can utilize the relationship of multiple modalities and extract more comprehensive information. The details of InnerGCN will be introduced in section \ref{Sec:section2.C}.

In addition, because of the instability of the traditional GAN framework and the unsmoothness of the training process, the results of generators are very poor. The generated data are quite different with the source data.
In view of above problems, the Balancer modules are add between the source data and the real input of discriminators, which can be seen in section \ref{Sec:section2.D} in details. The loss function of the model can not only calculate the differences in brain region level, but also take into account the overall differences. The details of Balancer modules can be seen in section \ref{Sec:section2.E}.

Based on above changes, BGGAN in this paper can realize the bidirectional mapping between structure and function at the same time by two generators and discriminators. The equations of BGGAN can be expressed in Eq.\ref{eq:eq2}.
\begin{equation}
	\begin{split}
		\mathcal{L}_{GAN}= \mathcal{L}_{SC2FC} + \mathcal{L}_{FC2SC} + \mathcal{L}_{recon}\times \lambda
	\end{split}
	\label{eq:eq2}
\end{equation}
Here, the $\mathcal{L}_{SC2FC}$ can be represented in Eq.\ref{eq:eq3}:
\begin{equation}
	\begin{split}
		\mathcal{L}_{SC2FC} = \mathcal{L}_{cla}^{sc}+\mathcal{L}_{cons}^{fc}+\mathcal{L}_{recons}^{sc}+\mathcal{L}_{dis}^{fc}+\mathcal{L}_{dis}^{s}\\
		= -\sum_{i=1}^{K}[y_ilog(\hat{y_i}+(1-y_i)\log(1-\hat{y_i}))]\\
		+ mse(real_{fc}, gen_{fc})
		+mse(real_{sc}, gen_{sc})\\
		+dis(gen_{fc})+dis(gen_{sc}).
	\end{split}
	\label{eq:eq3}
\end{equation}
and the $\mathcal{L}_{FC2SC}$ is same with the $\mathcal{L}_{SC2FC}$. $\mathcal{L}_{cla}^{sc}$ is the result of the classifier whose input is the generated data.  $\mathcal{L}_{cons}^{fc}$ is the construction loss of the generator from SC to FC and $\mathcal{L}_{recons}^{sc}$ is the reconstruction loss of the generator from FC to SC. $\mathcal{L}_{dis}^{fc}$ is the output of the discriminator whose input is the generated and real brain functional networks, while $\mathcal{L}_{dis}^{sc}$ is the output of the discriminator whose input is the regenerated and real brain structural networks.

\subsection{InnerGCN for extracting graphic features}
\label{Sec:section2.C}
Because of the topology of the brain, graph convolution networks have been widely used and achieved great results. Given a graph $G$ with an adjacency matrix $A \in \mathbb{R}^{N\times N}$, its diagonal degree matrix is represented as $D_{ii}=\sum_j A_{ij}$ and its normalized graph Laplacian can be denoted as $L = D^{-\frac{1}{2}}\hat{A}D^{-\frac{1}{2}} = I - D^{-\frac{1}{2}} A D^{-\frac{1}{2}}$, where $I$ is the identity matrix and $\hat{A}$ is the adjacency matrix $A$ with self-loop. And the  realization of graph convolution can be expressed as the Eq.\ref{eq:eq4}:
\begin{equation}
	\begin{split}
		X^{l+1} = \sigma (D^{-\frac{1}{2}}\hat{A}D^{-\frac{1}{2}}X^{l}W^{l}).
	\end{split}
	\label{eq:eq4}
\end{equation}
where $H^{l}$ represents the input of the layer $l$ and $W^{l}$ is the weight matrix of layer $l$.

But traditional GCN can only use the single modal or utilize one modal as the connection matrix and the other as the feature matrix. Based on these methods, information of the modal cannot be made full use. Tensor product, referring to a multiple dimensional array of numbers, is a powerful tool to analyze multidimensional data\cite{ng2011multirank,li2013multicomm,li2017low}. Inspired from \cite{huang2020mr,kilmer2011factorization,song2020robust}, we extend the input of graph convolution to high dimension space, which can utilize the features and connections of all modals at the same time. The mechanism of InnerGCN is illustrated in the following.

First, some variables are defined first. $\mathcal{A} \in \mathbb{R}^{R \times N \times N}$ is the networks' adjacency tensor, and features tensors of the brain networks can be defined as $X \in \mathbb{R}^{R \times N \times D}$. Here, $R$ is the modalities' number of the input, $N$ is the brain regions' number used in this paper and $D$ is the dimension of the features' matrix. Due to the need of the formula below, we define $A(i)$ as $A(i) = A(i, :, :)$. And then the InnerGCN is as following:
\begin{equation}
\begin{split}
    x \star   g = U^H \diamond ((U \diamond x) \odot (U \diamond g)).
\end{split}
\label{eq:eq5}
\end{equation}
where $x$ is the features' tensor of the input and $g$ is the convolution kernel.

Next, we will introduce how to get $U$ in next steps:
1) get the Laplace tensor $L$, where $L(i) = I-D(i)^{-1/2}A(i)D(i)^{-1/2}$; 2) get the $L_{dft}$ from the $L$ in the Fourier form; 3) get the singular values matrix of each dimension of the tensor $[u(i), s(i)] = EVD(L_{dft}(i))$; 4) map the data in the Fourier data back to the original space $U= IDFT(u), \Lambda=IDFT(s)$. $U^H$ is the Hermitian matrix of $U$. $U \diamond x$ is the $U$ of Fourier form, and same as convolution filter $g$.
The third-order tensor can be considered as matrix with each element to be a vector. Same with matrix multiplication based on vector multiplication, matrix-matrix interaction is realized through circular convolution based on matrix multiplication. We defined $A, B$ as $A \in \mathbb{R}^{n_1 \times n_2 \times 3}$ and $B \in \mathbb{R}^{n_2 \times n_3  \times 3}$ and fold1(...), fold2(...), unfold(...) in Eq (\ref{eq:eq6}-\ref{eq:eq9}):
\begin{equation}
    \begin{split}
        fold1(A) = \left[
            \begin{array}{cc}
                 A_1 \\
                 A_2 \\
                 A_3 \\
            \end{array}
        \right]
    \end{split}
    \label{eq:eq6}
\end{equation}
\begin{equation}
    \begin{split}
        fold2(B) = \left[
            \begin{array}{ccc}
                 B_{11} & B_{12} & B_{13} \\
                 B_{21} & B_{22} & B_{23} \\
                 B_{31} & B_{32} & B_{33} \\
            \end{array}
        \right]
    \end{split}
    \label{eq:eq7}
\end{equation}
\begin{equation}
    \begin{split}
        unfold(\left[
            \begin{array}{ccc}
                 B_{11} & B_{12} & B_{13} \\
                 B_{21} & B_{22} & B_{23} \\
                 B_{31} & B_{32} & B_{33} \\
            \end{array}
        \right]) = B
    \end{split}
    \label{eq:eq8}
\end{equation}
\begin{equation}
    \begin{split}
    A \diamond B = \mathcal{F}^{-1}(unfold(fold1(\mathcal{F}(A)) \times fold2(\mathcal{F}(B))))
    \end{split}
    \label{eq:eq9}
\end{equation}

In short, the process of InnerGCN is to Fourier transform the node features and the convolution kernel into the Fourier domain, and then let the two matrix multiply slice-by-slice in the Fourier domain, and finally inverse Fourier transform the result into the original space. It is circular convolution that makes the InnerGCN can nicely utilize relation correlations.

\subsection{Balancer for smoothing training process of BGGAN}
\label{Sec:section2.D}
The basic idea of GAN is to train a nonlinear function, which can map the brain network from one domain to the other. To achieve this goal, the discriminator is trained to measure the distance of distribution between source and target data. The feedback of generators and discriminators can make generators learn mapping function from source domain to target domain.

In this paper, structural and functional brain networks are too distant from each other. Even if there is a similar mapping function between structural and functional domain, generators can not learn mapping function very well and details of the output can't be generated. To decrease the distance between structural and functional domain, the Balancer is added between the real data of discriminator and the source data. The main idea of Balancer is to blur the real data of the discriminator and create a hyper-parameter to control the blurring level.
As the epoch of training iterations increases, the generators of BGGAN gradually learn mapping function from the source domain to the target domain. Besides, since the ability of the discriminator is limited by the generator and the epoch of training epoch, the generators can gradually learn the details of the source data, making the output of the generator with Balancer much better than that without Balancer.

A neural network is utilized as Balancer, which is comprised of several convolution networks, a step connection combining the high order information and the low order information from the Balancer. Its loss function contains two parts: an adversarial loss $\mathcal{L}_B^{adv}$ and a reconstruction loss $\mathcal{L}_B^{recon}$.
The function of the adversarial loss is to try fool the discriminator and slow down the discriminators' learning process. The reconstruction loss is to limit the output similar to real sample and avoid the discriminator learn incorrect distribution too much.
The Balancer is trained as following loss:
\begin{equation}
	\begin{split}
		\mathcal{L}_L = \mathcal{L}_L^{adv} +\mathcal{L}_L^{recon} = -\mathcal{L}_D + \mathcal{L}_L^{recon}\\ = -D(\mathcal{X}) + ||\mathcal{X}-\mathcal{Y}||^2_2.
	\end{split}
	\label{eq:eq11}
\end{equation}
where $\mathcal{X}$ is the input of Balancer, $\mathcal{Y}$ is the output of Balancer and $\mathcal{D}$ is a discriminator in the model.

To accelerate the training process, $\lambda$ is added in the loss function of Balancer. To have a smooth training procedure, $\lambda$ should be decreased gradually and the curve of $\lambda$'s values should be relatively smooth curve. After a certain training epochs, generators and discriminators have learned the relationship between the input and the output, Balancer should be deleted and make the model learn the details of the input. Considering above situation, the $\lambda$ is designed as shown in Eq(8):
\begin{equation}
	\begin{split}
		\lambda = \left\{
		\begin{matrix}
			e^{-0.01t} & t \le K  \\ 
			0 & t > K
		\end{matrix}	
		\right.
	\end{split}
	\label{eq:eq12}
\end{equation}

The $\lambda_p$ is obtained from the training procedure to normalize the $\lambda$ to between 1 and 2. Therefore, the overall loss of Balancer should be:
\begin{equation}
	\begin{split}
		\mathcal{L}_L = \mathcal{L}_L^{adv} +\mathcal{L}_L^{recon} = -\lambda\mathcal{L}_D + \mathcal{L}_L^{recon}.
	\end{split}
	\label{eq:eq13}
\end{equation}

In order to achieve above goals, we will design a new module named Balancer.
It is mainly composed of convolution layers and ReLU activation layers. Because the input contains structural and functional information, the output of the balancer will contain both structural and functional information. Based on this, generators can learn structural and functional information at the same time.

\subsection{Loss function for model}
\label{Sec:section2.E}
The loss function are composed by the following modules: 1) Classifier; 2) Generators; 3) Discriminators; 4) Balancers.

The overall loss function of the framework can be expressed as the following form:
\begin{equation}
	\begin{split}
		\mathcal{L} = \mathcal{L}_{gan} + \mathcal{L}_{cons} + \mathcal{L}_{recon} + \mathcal{L}_{inden} + \mathcal{L}_{cla}.
	\end{split}
	\label{eq:eq14}
\end{equation}
The $\mathcal{L}_{gan}$ is the output of the discriminator through calculating the distribution gap between generated and the source brain connections. The $\mathcal{L}_{cons}$ is the difference between generated data and the real data of source domain. The $\mathcal{L}_{recon}$ is the difference between reversely generated results and the real data of same domain. By means of calculating the generated results and real brain matrices of the original domain, the gap can be obtained and expressed as $\mathcal{L}_{index}$. The $\mathcal{L}_{cla}$ is obtained by the classifier to reflect the degree to which different types of features are mapped to different areas in the latent space.

Based on loss function \ref{eq:eq14}, the model is gradually improved after each training epoch. Both the generated results and the classification results are optimized at the same time.

\begin{algorithm}[t]
\label{alg:al1}
\caption{Bidirectional Inference algorithm on BGGAN}
\hspace*{0.02in} {\bf Input:}
each modalities' feature information $F$, adjacency matrix $G$, subjects' label $y$, maximum iterative number $EPOCH$ and hyperparameters $\lambda_{sc}$, $\lambda_{fc}$\\
\hspace*{0.02in} {\bf Output:}
structural and functional generated connections $G'$ and model's parameters $\Theta$
\begin{algorithmic}[1]
\State initialize model's parameters $\Theta$, hyperparameters $\lambda_{sc}$, $\lambda_{fc}$ and $EPOCH$ = k, epoch = 0
\For{epoch $<$ EPOCH}
	\State epoch $\leftarrow$ epoch + 1
	
	\State update $\lambda$: $\lambda=e^{-0.01 \times epoch}$ before epoch $<$ k

	\State combine structural and functional features and
	\Statex $\quad\quad\quad$ connectivity matrices($F_{sc}$, $E_{sc}$), ($F_{fc}$, $E_{fc}$)
	
	\State compute the latent feature of structure and function
	\Statex $\quad\quad\quad$ ($F'_{sc}$, $F'_{fc}$) by Equation: $U^T(U x \odot U g)$
	
	\State classify the subjects' category based on ($F'_{sc}$, $F'_{fc}$)
	
	\State generate the structural and functional adjacency
	\Statex $\quad\quad\quad$ matrices: $E' = F' \times T(F')$
	
	\State discriminate the generated structural and functional
	\Statex $\quad\quad\quad$ adjacency matrices:
	\Statex $\quad\quad\quad$ $\mathcal{L}'_{gan} = Loss'_{gen}+Loss'_{dis}$
	
	\State recompute the latent feature of structure and function
	\Statex $\quad\quad\quad$ ($F''_{sc}$, $F''_{fc}$) by Equation: $U^T(U x \odot U g)$
	
	\State regenerate the structural and functional adjacency
	\Statex $\quad\quad\quad$ matrices : $E'' = F'' \times T(F'')$
	
	\State rediscriminate the generated structural and
	\Statex $\quad\quad\quad$ functional adjacency matrices:
	\Statex $\quad\quad\quad$
	$\mathcal{L}''_{gan} = Loss''_{gen}+Loss''_{dis}$
	
	\State compute the loss in the whole model:
	\Statex $\quad\quad\quad$ $\mathcal{L}_{cons} = ||E, E'||_2$
	\Statex $\quad\quad\quad$ $\mathcal{L}_{recon} = ||E', E''||_2$
	\Statex $\quad\quad\quad$ $\mathcal{L}_{inden} = ||E, E''||_2$
	\Statex $\quad\quad\quad$ $\mathcal{L}_{gan}=L'_{gan}+\mathcal{L}''_{gan}$
	\Statex $\quad\quad\quad$ $\mathcal{L} = \mathcal{L}_{gan} + \mathcal{L}_{cons} + \mathcal{L}_{recon} + \mathcal{L}_{inden} + \mathcal{L}_{cla}$
	
	\State update the parameters of the model:
	\Statex $\quad\quad\quad$ $\theta_{module} = \theta_{module} - \alpha \bigtriangledown_{\theta_{module}}L_{module}$
\EndFor
\State
\Return generated brain functional and structural adjacency matrices
\end{algorithmic}
\end{algorithm}

\section{Experiments}
\label{Sec:Experiments}
\subsection{Datasets and Experimental Settings}

In this subsection, a series of experiments are conducted based on the Alzheimer’s Disease Neuroimaging Initiative (ADNI) dataset \cite{adding_300}. In our experiments, we used 80\% of the subjects as the training dataset and 20\% of the subjects as the testing dataset. All experimental results are based on the testing dataset. The details of the datasets used in this paper are presented in Tab.\ref{Tab:Tab3-1}, which provides the number of subjects, gender distribution, mean age, and mean weight (with standard deviations) across five categories: Normal Control (NC), Significant Memory Concern (SMC), Early Mild Cognitive Impairment (EMCI), Late Mild Cognitive Impairment (LMCI), and Alzheimer’s Disease (AD). This detailed summary allows for a clear understanding of the dataset composition and demographic distribution used in the experiments.

\begin{table}[h]
	\centering
	\caption{The situation of data used in the experiment}
	\begin{tabular}{p{1cm} | p{1cm} p{1cm} p{1cm} p{1cm} p{1cm}}
		\toprule
		Category & NC & SMC & EMCI & LMCI & AD \\
		\midrule
		Number & 153 & 94 & 135 & 63 & 64 \\
		Gender & 81/72 & 35/59 & 86/49 & 35/28 & 39/25 \\
		Age & 73.5$\pm$8.0 & 76.2$\pm$5.1 & 75.7$\pm$6.8 & 75.8$\pm$6.1 & 74.7$\pm$7.6 \\
		Weight & 74.7$\pm$16.5 & 79.7$\pm$16.1 & 80.0$\pm$13.0 & 76.9$\pm$16.6 & 74.2$\pm$14.1 \\
		\bottomrule
	\end{tabular}
	\begin{tablenotes}
		\footnotesize
		\item The values are denoted as mean $\pm$ standard deviation. Categories contains Normal Control(NC), Significant Memory Concern(SMC), Early Mild Cognitive Impairment(EMCI), Late Mild Cognitive Impairment(LMCI) and Alzheimer's Disease(AD).
	\end{tablenotes}
	\label{Tab:Tab3-1}
\end{table}

\subsection{The effectiveness of InnerGCN}
In this subsection, several comparison experiments are conducted to evaluate the effectiveness of InnerGCN.
The hyperparameters of the model used in this section are defined in Tab.\ref{Tab:tab3-2}, which outlines the optimal parameter settings, including the optimizer type, learning rates for generators and discriminators, batch size, stop strategy, number of regions of interest (ROIs), and the modalities used (sMRI, DTI, and fMRI). These parameters were selected based on preliminary experiments and aim to ensure the most robust and accurate model performance in the conducted experiments.

\begin{table}[h]
	\centering
	\caption{The optimal parameters settings of the Model }
	\begin{tabular}{l|l}
		\toprule
		Parameter Name & Parameter Value\\
		\midrule
		Optimizer & Adaptive Moment Estimation(Adam) \\
		Learning rate of Generators &  0.0005 \\
		Learning rate of Discriminators &  0.0005 \\
		Batch size & 8 \\
		Stop Strategy & the early stop strategy \\
		Number of ROIs & 90 \\
		Types of Modality & sMRI,DTI,fMRI \\
		\bottomrule
	\end{tabular}
	\label{Tab:tab3-2}
\end{table}
Based on the above settings
, three other comparison graph convolution methods, including GCN\cite{kipf2016semi}, GAE\cite{kipf2016variational} and GAT\cite{velivckovic2017graph} methods, are used to illustrate the effectiveness of InnerGCN. The classification indicators are used to assess these graph convolution methods' performance.

Multiple sets of comparison experiments are used to validate the results achieved by InnerGCN in multiple modalities. Classification experiments are performed using multiple modalities' fusion of fMRI and DTI data. For methods GCN, GAE and GAT, the node features are derived from the time series of fMRI, while the edge connection matrix are from the data of DTI. In addition, the InnerGCN takes both node features and edge connectivity information from fMRI and DTI data as model's input.
The classification indicators $ACC$, $Recall$, $Precision$ and $F1\_score$ are used to evaluate the performance advantages and disadvantages with these different classification methods.

\begin{figure}[htb]
	\centering
	\includegraphics[width=0.8\linewidth]{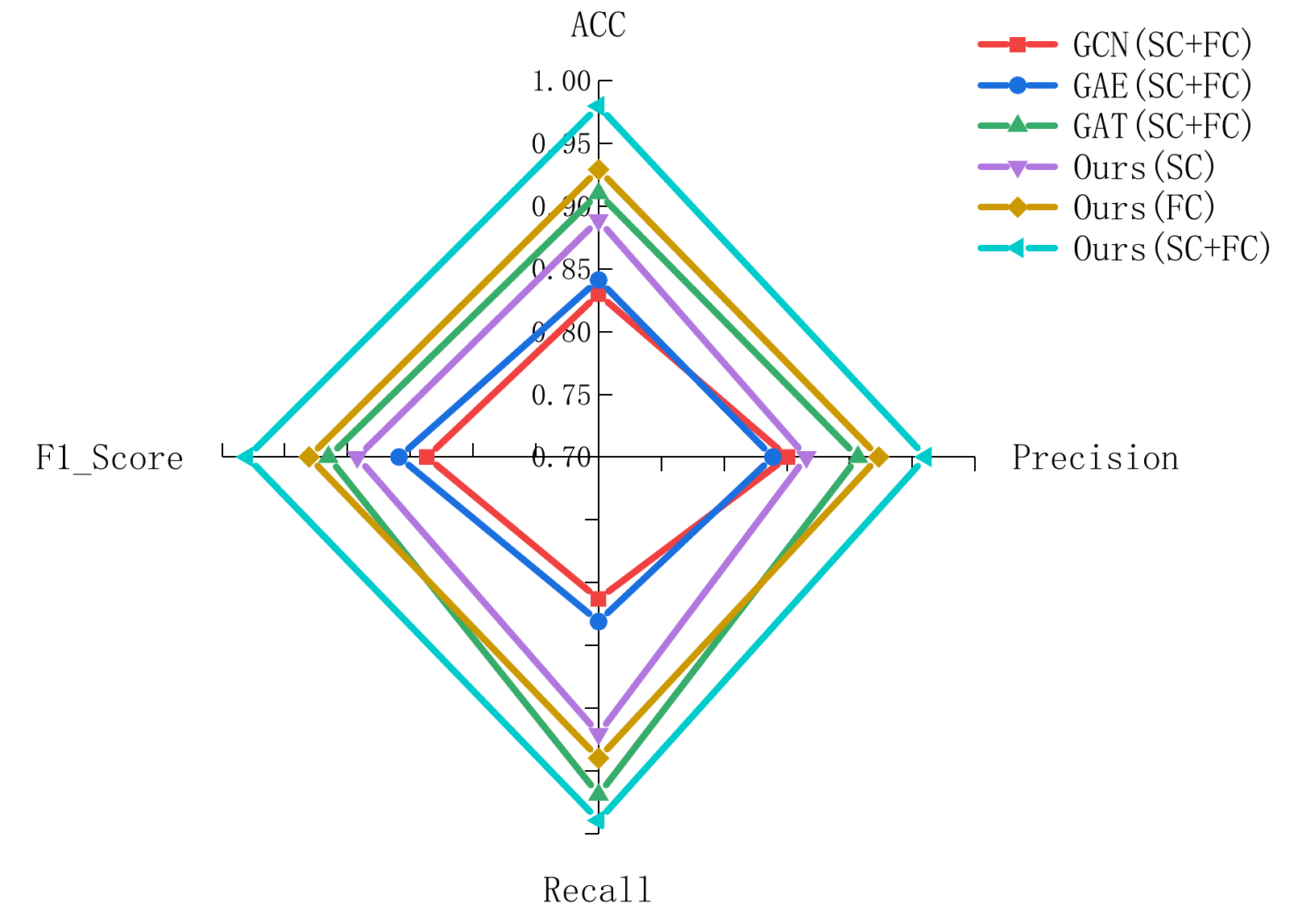}
	\caption{Classification performance based on different graph convolution methods. GCN, GAE and GAT set the brain structural information as the edges of graph and the brain functional information as the node features of graph. The classification result based on our method reaches best.}
	\label{fig:Fig4}
\end{figure}

As shown in Fig.\ref{fig:Fig4}, both GAT and InnerGCN show a great improvement in classification results
compared with other graph convolution methods. However, InnerGCN can better utilize the complementary information between these two multiple modal data, which improves the classification accuracy greatly and proves the validity of the InnerGCN.

\subsection{The effectiveness of the Balancer}
In fact, generative adversarial networks are subject to significant instabilities. In the models like ours, discriminators tend to reach the optimal case more quickly than generators, which leaves a great performance divide between discriminators and generators in the same model.

To tackle this problem, the training speed of the discriminator needs to slow down and the gap between source domain and target domain should be reduced. In other words, retarding the optimization process of the discriminator makes the generator's training more stable.

To weight these important factors above, a new module named Balancer is proposed and added between source data and discriminators. Before adding the module, the source data is directly inputted to the discriminator. In order to retard the learning speed of the discriminator and try to improve the performance of the generator step by step, the source domain and target domain are integrated together into the input of the discriminator until multiple epochs pasted.

\begin{figure}[htb]
	\centering
	\subfigure[]{
        \begin{minipage}[t]{0.95\linewidth}
            \centering
            \includegraphics[width=0.95\linewidth]{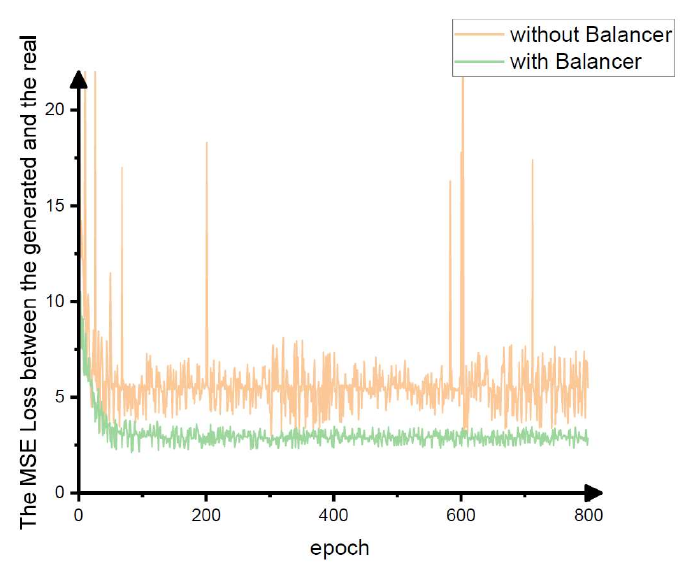}
            \label{fig:5_1}
        \end{minipage}
    }
	\subfigure[]{
        \begin{minipage}[t]{\linewidth}
            \centering
            \includegraphics[width=0.95\linewidth]{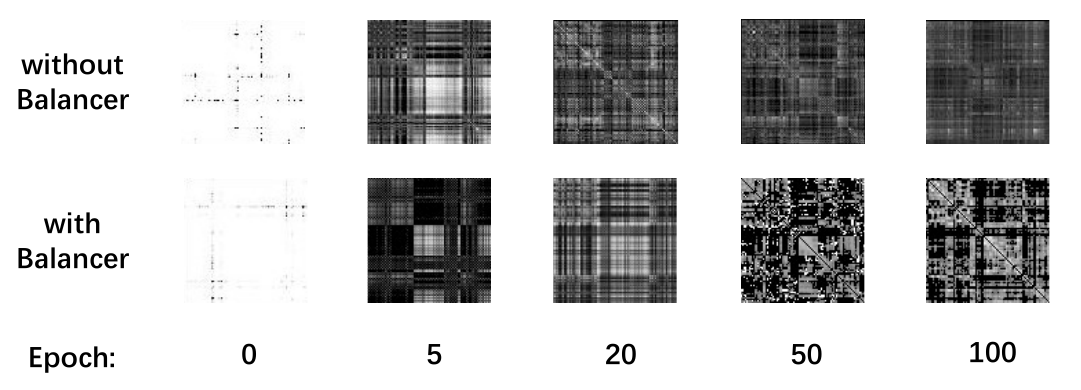}
            \label{fig:5_2}
        \end{minipage}
    }
    \caption{(a) shows the MSE loss between the target and the source domain connections. The orange curve is the loss outputted by the generator without module Balancer, while the green curve is the loss with module Balancer. (b) shows the generated structural connections from the functional generator. The top five pictures are the results from the generator without module Balancer and the bottom five are the results from the generator with module Balancer.}
    \label{fig:Fig5}
\end{figure}
Here we conduct two sets of comparison experiments, one without Balancer and one with Balancer. From Fig.\ref{fig:5_1}, it is clear that there is a large difference in the loss of the functional generator in two cases. The generator with module Balancer has better training results and the training process is more stable, while the generator without module Balancer can reach a better level faster at the beginning of training. Based on the model without module Balancer, the discriminator reaches a better level soon. As a result, although discriminators' grade is more accurate, the generators' next training will be more difficult instead. When the training process reaches a certain level, the generator has learned some distributions. The training of the discriminator should also change with the training process of the generator. Just as shown in Fig.\ref{fig:5_2}, the top five pictures show generated brain functional connections without module Balancer and the bottom five pictures are the generated brain functional connections with the module Balancer. Although effective results can be achieved regardless of whether the module Balancer is used or not, the results tell us that the generated results with the Balancer get more realistic details as well.

\subsection{The connection number changes from NC and AD in structural and functional domain}

Although many articles have stated that structure is the basis of function, there is actually a huge difference between structure and function in terms of brain analysis performance.

In this section, we will discuss differences of brain structure and function in terms of the number of brain connections. In order to reflect the relationships between brain structure and function, we compared the brain connection number changes in structure and function from NC to AD.

\begin{figure}[htb]
	\centering
	\subfigure[]{
        \begin{minipage}[t]{1.0\linewidth}
            \centering
            \includegraphics[width=0.9\linewidth]{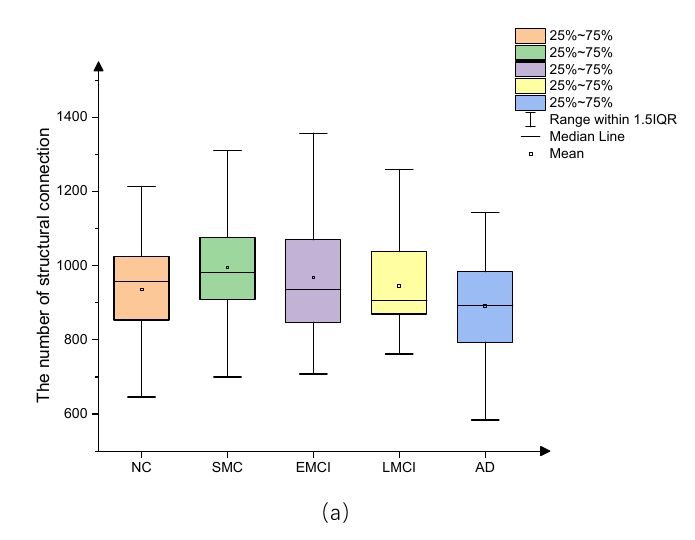}
			\label{fig:6_1}
        \end{minipage}
    }
	\subfigure[]{
        \begin{minipage}[t]{1.0\linewidth}
            \centering
            \includegraphics[width=0.9\linewidth]{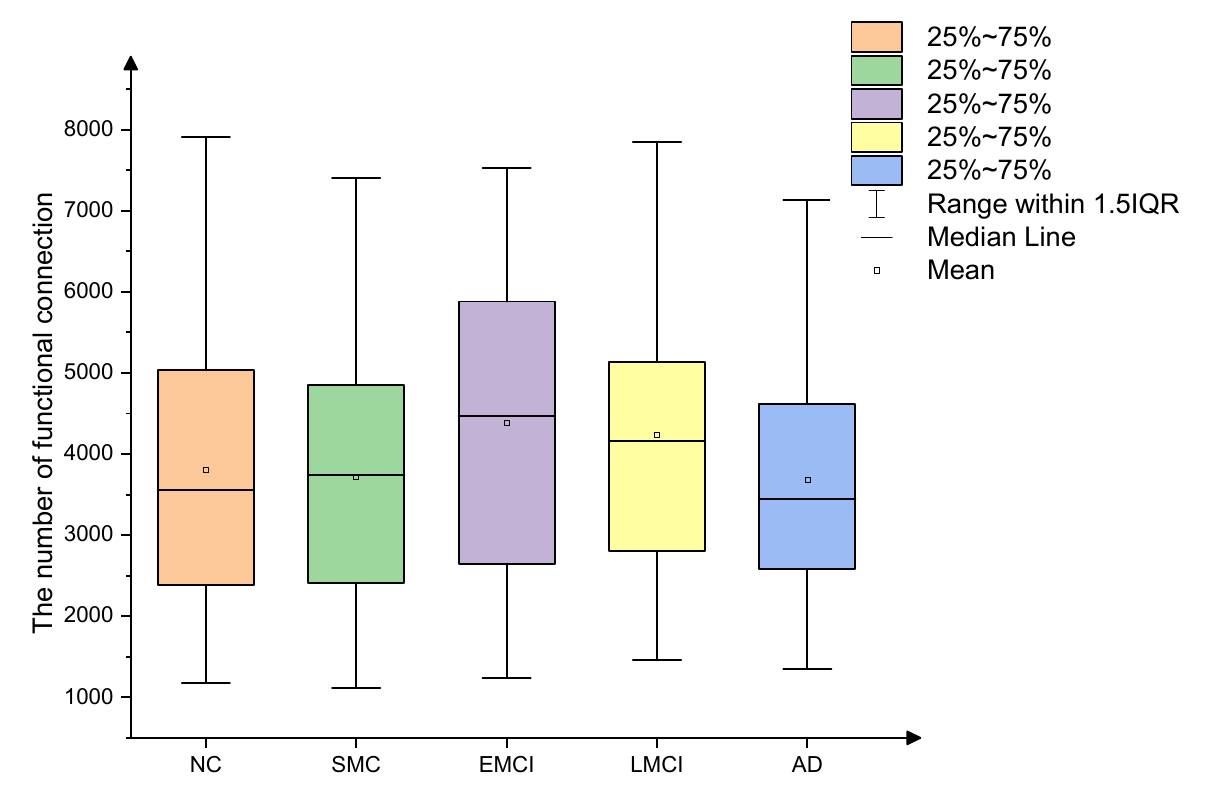}
            \label{fig:6_2}
        \end{minipage}
    }
	\caption{Statistical analysis for structural and functional brain connections under each category. The figure (a) shows the number trend of structural connections and the figure (b) shows the number trend of functional connections. }
	\label{fig:Fig6}
\end{figure}

Then structural and functional brain connections are compared in the large aspect of the same category. 60 subjects in each category are identified for generating structural and functional brain connections. Then the number of structural connections and the number of functional connections are counted. The statistical results are shown in Fig.\ref{fig:Fig6}. As shown in Fig.\ref{fig:6_1} and Fig.\ref{fig:6_2}, the number of structural brain connections increased first and then decreased. This situation is same as the functional brain connections.
This also confirms that Alzheimer's disease does have a strong relationship with the number of brain structural and functional connections. The difference is that the start of the decline in the number of structural brain connections occurs at SMC, whereas the decline in the number of functional brain connections occurs at the EMCI stage, a later stage of SMC. Maybe, the greater coordination of the functional brain domains and the relatively more stable structure of the brain allowed the growth of functional brain connectivity to not begin to decline until the EMCI stage.

\subsection{The brain adjacency changes from NC to AD}
\begin{figure*}[htb]
	\centering
	\includegraphics[width=0.8\linewidth]{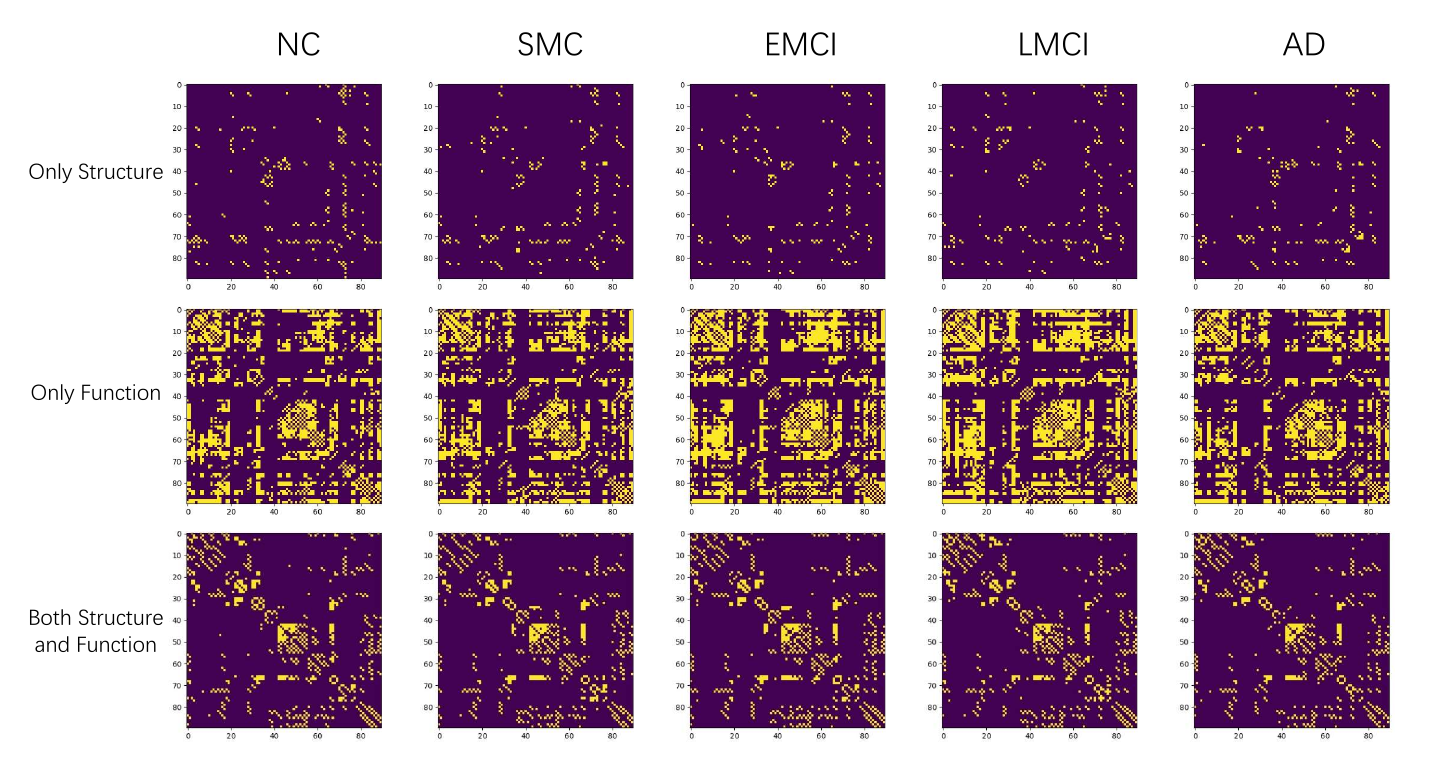}
	\caption{Similarities and differences between brain structural connections and brain functional connections in each category. The brighter the plot, the higher the probability of presence of brain connections. The yellow areas indicate a higher likelihood of brain connections, while the rest of the purple colors indicate that brain connections are less likely to be present. Top five figures show brain connections where structural connections usually exist but functional connections do not. The five figures in the middle show the brain connections that have a high probability of function but rarely structure. And the bottom five show brain connections that both exist multiple times in structure and function.}
	\label{fig:Fig7}
\end{figure*}

Then, we compared the differences in brain structure and function under the perspective of brain regions. Here, we performed multiple comparison experiments using the averaging method, including \{NC vs. SMC, NC vs. EMCI, NC vs. LMCI, NC vs. AD\}. Fig.\ref{fig:Fig7} shows the comparative results of these experiments.

The results in Fig.\ref{fig:Fig7} are calculated in Independent Samples t-Test method based on brain generated structural and functional connections which are synthesized by trained generators above.
The top five diagrams show brain connections where structure is often present but function is infrequent; the middle five diagrams show brain connections where function occurs many times but structure rarely; and the bottom five diagrams show brain connections where both structure and function occur with high probability. The brighter the plot, the higher the probability of the presence of brain connections. There are similarities in the data distribution in each row as a whole, but there are huge differences in some regions.

Based on above comparison experiments, functional brain connections tend to occur between brain regions that are directly connected to brain structures, and conversely, brain regions with functional brain connections are not necessarily directly connected to each other by structural brain connections. However, the fact is that the number of brain functional connections is much larger than that of brain structural connections. There are also brain functional connections between many brain regions that do not have brain structural connections. We can reasonably speculate that functional brain connections may be the result of multiple brain regions acting together to achieve the corresponding functions.

Overall, there is a one-to-one correspondence between structure and function in some connections, but on a larger scale there is not a one-to-one correspondence between structural and functional brain connections. Brain structure is the basis of brain function, demonstrating that there are complex relationships in the brain that allow information transfer between brain regions without structural connections to perform the corresponding functions.

\subsection{Differences of brain adjacency matrices in different categories compared with NC}
In this section, 60 subjects' data from each category are randomly selected in this experiment. The brain structural and functional adjacency matrices based on the Independent Samples t-Test method is conducted to obtain the difference in haunting relationships between patients and normal subjects.

\begin{figure*}[htb]
	\centering
	\includegraphics[width=0.9\linewidth]{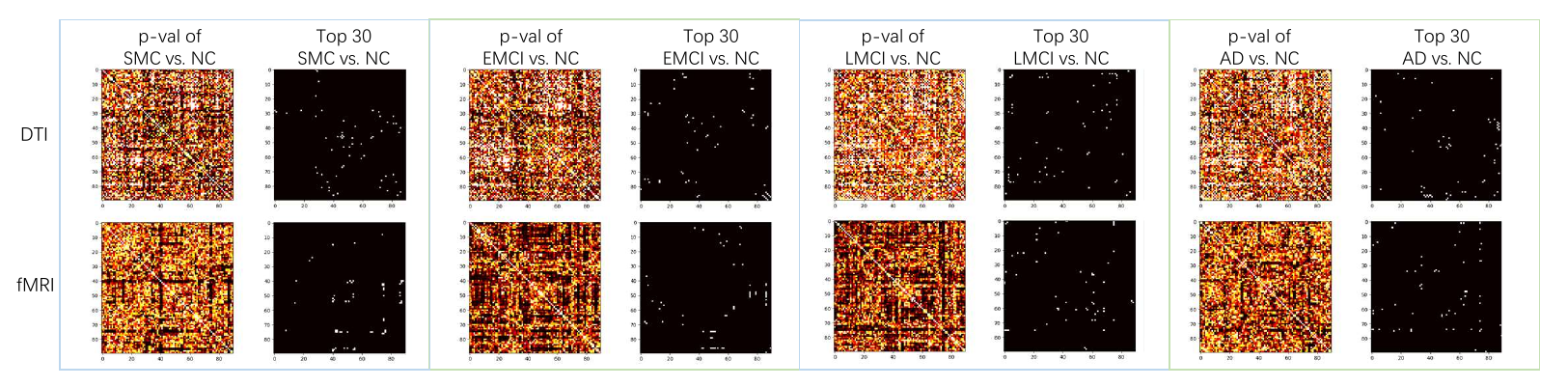}
	\caption{The p-value results based on the Independent Samples t-Test of structure and function under adjacent categories compared with normal control. The top four images are structural comparison results and the bottom four images are functional comparison results.}
	\label{fig:Fig9}
\end{figure*}

According to Fig.\ref{fig:Fig9}, the more yellow the color, the greater the p-value, which means that adjacent categories will have brain connections present between these two brain regions. It can be clearly seen that the structural changes are less than the functional during the progression of the disease. But the changes of structure and function are almost in similar adjacent brain regions.

\begin{table}[]
\caption{Top 30 abnormal brain connections obtained using Independent Samples t-Test experiments compared to normal subjects. The numbers in this table represent brain area number.}
\begin{tabular}{cc|cc|cc|cc}
\toprule
\multicolumn{2}{l}{\small{\textbf{SMC vs. NC}}}        & \multicolumn{2}{l}{\small{\textbf{EMCI vs. NC}}}       & \multicolumn{2}{l}{\small{\textbf{LMCI vs. NC}}}                               & \multicolumn{2}{l}{\small{\textbf{AD vs. NC}}}         \\ \midrule
\small{SC}                           & \small{FC}  & \small{SC}                           & \small{FC} & \small{SC}                           & \small{FC}                          & \small{SC}                           & \small{FC}  \\ \midrule
\cellcolor[HTML]{E2EFDA}40-56 & 56-88 & 6-31                          & 50-87 & \cellcolor[HTML]{E2EFDA}40-56 & 37-61                         & \cellcolor[HTML]{38FFF8}85-87 & 28-76 \\ \hline
46-47                         & 55-88 & 4-30                          & 49-76 & 6-27                          & 6-25                          & 52-86                         & 69-75 \\ \hline
73-82                         & 41-88 & 46-47                         & 50-76 & 6-31                          & 14-23                         & 37-82                         & 20-75 \\ \hline
47-48                         & 76-77 & 14-78                         & 58-87 & 6-25                          & 3-76                          & 65-85                         & 28-75 \\ \hline
\cellcolor[HTML]{38FFF8}85-87 & 44-75 & 29-71                         & 87-90 & 12-62                         & 63-65                         & 37-87                         & 2-17  \\ \hline
47-68                         & 68-76 & 86-88                         & 50-80 & 4-30                          & \cellcolor[HTML]{FD6864}30-72 & 7-83                          & 49-65 \\ \hline
19-29                         & 43-76 & \cellcolor[HTML]{38FFF8}85-87 & 5-70  & 5-6                           & 57-61                         & 39-87                         & 8-61  \\ \hline
60-62                         & 51-76 & 38-43                         & 5-38  & 4-73                          & 2-76                          & 53-89                         & 34-45 \\ \hline
\cellcolor[HTML]{FCFF2F}52-54 & 41-53 & \cellcolor[HTML]{FD6864}30-72 & 60-65 & 4-75                          & 23-29                         & \cellcolor[HTML]{9698ED}41-87 & 75-76 \\ \hline
48-54                         & 41-55 & 6-32                          & 51-76 & 38-55                         & 69-76                         & 63-85                         & 2-75  \\ \hline
38-82                         & 43-75 & 88-90                         & 57-87 & 41-71                         & 36-53                         & 7-23                          & 6-75  \\ \hline
33-73                         & 44-76 & \cellcolor[HTML]{9698ED}41-87 & 44-87 & 77-78                         & 9-38                          & 48-54                         & 79-90 \\ \hline
2-30                          & 47-76 & 14-77                         & 49-87 & 23-51                         & 63-69                         & \cellcolor[HTML]{FCFF2F}52-54 & 29-64 \\ \hline
44-47                         & 75-77 & 8-24                          & 30-71 & 58-59                         & 56-88                         & 38-88                         & 49-66 \\ \hline
9-31                          & 67-76 & 34-77                         & 16-69 & \cellcolor[HTML]{38FFF8}85-87 & 22-76                         & 35-85                         & 16-76 \\ \hline
44-86                         & 46-76 & \cellcolor[HTML]{E2EFDA}40-56 & 14-60 & 25-71                         & 38-62                         & 7-31                          & 31-35 \\ \hline
48-67                         & 41-89 & 6-76                          & 43-87 & 30-62                         & 50-62                         & 47-50                         & 52-75 \\ \hline
36-48                         & 16-41 & 4-76                          & 59-65 & 30-31                         & 38-61                         & 53-55                         & 54-62 \\ \hline
38-84                         & 42-88 & 32-78                         & 54-87 & 4-68                          & 43-57                         & 77-78                         & 5-15  \\ \hline
\cellcolor[HTML]{FD6864}30-72 & 50-55 & 14-24                         & 52-76 & 7-78                          & 58-62                         & 55-87                         & 17-58 \\ \hline
1-29                          & 25-27 & 32-33                         & 4-69  & 59-82                         & 1-76                          & 40-88                         & 2-64  \\ \hline
\cellcolor[HTML]{9698ED}41-87 & 51-87 & 80-86                         & 52-87 & 22-78                         & 2-36                          & 35-81                         & 2-62  \\ \hline
33-82                         & 56-87 & 1-83                          & 6-44  & 2-86                          & 43-69                         & 42-88                         & 72-89 \\ \hline
29-68                         & 15-56 & 34-37                         & 44-76 & 10-73                         & 2-3                           & 29-34                         & 35-44 \\ \hline
29-33                         & 52-55 & 40-88                         & 9-55  & 30-73                         & 57-89                         & 52-90                         & 58-64 \\ \hline
43-74                         & 43-87 & 39-78                         & 17-58 & 35-74                         & 61-69                         & 72-88                         & 1-75  \\ \hline
78-84                         & 9-75  & 87-89                         & 30-65 & 4-77                          & 4-76                          & 29-57                         & 35-47 \\ \hline
37-78                         & 42-54 & \cellcolor[HTML]{FCFF2F}52-54 & 49-80 & 5-78                          & 1-43                          & 60-78                         & 57-75 \\ \hline
52-60                         & 42-51 & 33-51                         & 15-56 & 29-68                         & 10-23                         & 10-42                         & 52-65 \\ \hline
74-77                         & 49-87 & 7-78                          & 51-87 & 25-78                         & 1-44                          & 4-19                          & 25-35 \\ \bottomrule
\end{tabular}
\label{table:Tab3-3}
\end{table}

Based on above four comparison results, we further counted the first 30 abnormal brain connections lower than 0.05 in these experiments compared to normal subjects, and the results are shown in the Tab.\ref{table:Tab3-3}. From the table, a clear phenomenon can be found:
there were multiple abnormal brain connections in the structural experiments, and these brain connections appeared many times in these four groups of experiments, while there were no such abnormal brain connections in the functional experiments, or at most in two groups of experiments. This proves that the deterioration of brain structure is gradual, but through its coordination mechanism, the brain can skip the original abnormal brain connections, use other connections to achieve the same task, which are different from the brain connections of NC.

\begin{figure*}[htbp]
	\centering
 	\includegraphics[width=0.75\linewidth]{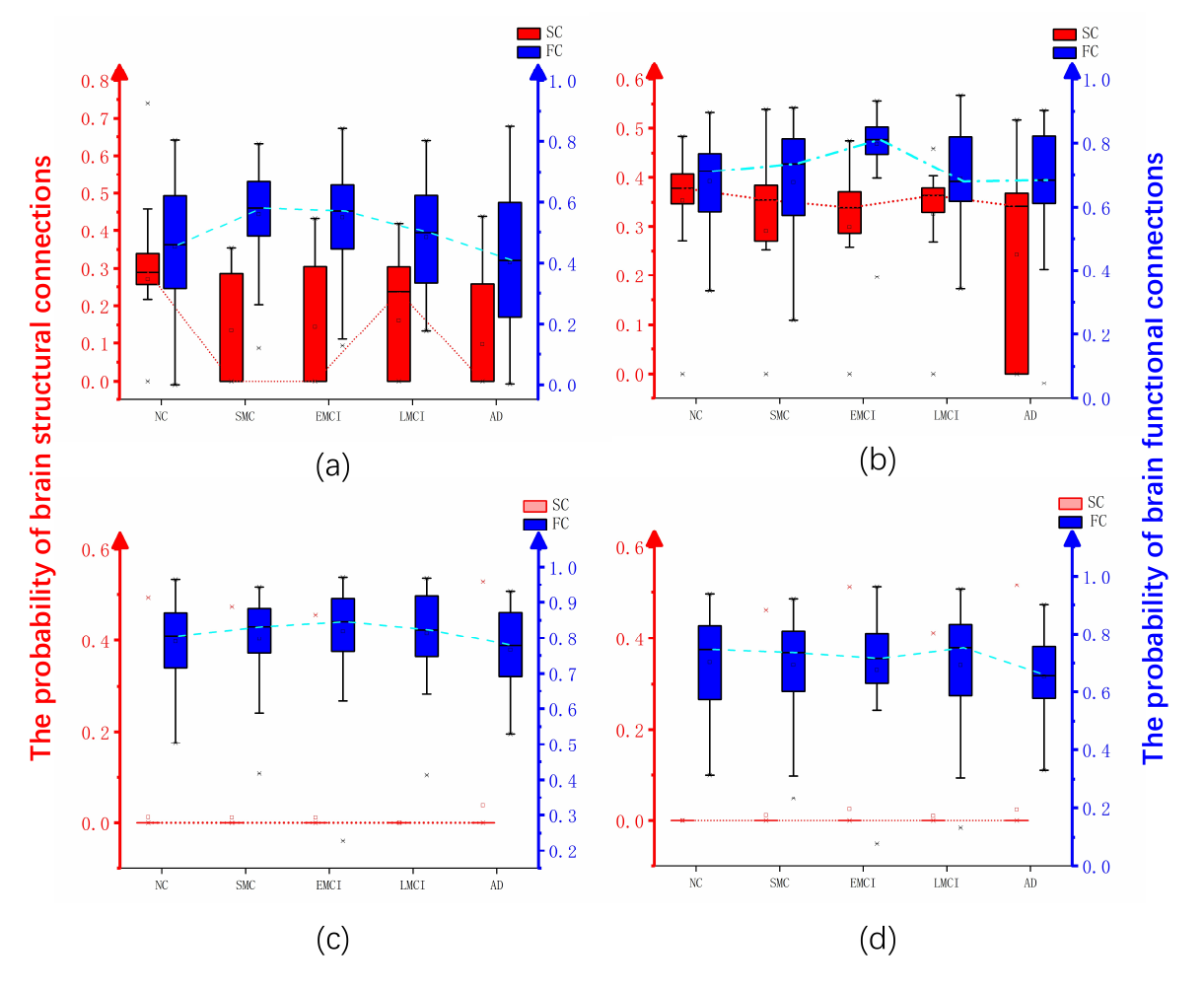}
	\caption{(a) shows the brain abnormal connection between brain region 85 and 87, (b) shows the brain abnormal connection between brain region 52 and 54; while (c) shows the brain normal connection between brain region 83 and 84, (d) is the brain normal connection between brain region 85 and 86.}
	\label{fig:Fig10}
\end{figure*}

\begin{figure}[h]
	\centering
	\includegraphics[width=0.9\linewidth]{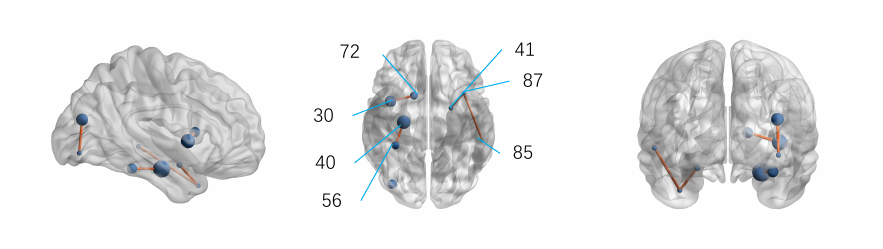}
	\caption{Abnormal brain structural and functional connections that are present in these comparison experiments.}
	\label{fig:Fig11}
\end{figure}

\begin{figure}[htb]
	\centering
	\includegraphics[width=0.9\linewidth]{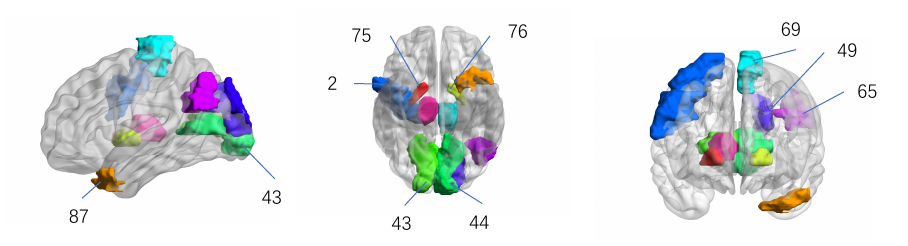}
	\caption{Abnormal brain structural and functional regions that are present multiple times in these comparison experiments.}
	\label{fig:Fig12}
\end{figure}

By comparing these first 30 brain connections, just shown in Tab.\ref{table:Tab3-3}, it was found that the structural data is relatively stable, such as the brain connections between brain region Occipital\_Mid\_R and Occipital\_Inf\_R or brain region Temporal\_Mid\_L and Temporal\_Pole\_Mid\_L, appeared many times in these four groups of experiments. For this reason, we utilized 60 subjects and counted the connection strength of their brain connections between brain region Temporal\_Mid\_L and Temporal\_Pole\_Mid\_L, and between brain region Occipital\_Mid\_R and Occipital\_Inf\_R. The statistical result is shown in Fig.\ref{fig:Fig10}. Fig.\ref{fig:Fig10}(a) and Fig.\ref{fig:Fig10}(b) are the abnormal brain connections' results, while Fig.\ref{fig:Fig10}(c) and Fig.\ref{fig:Fig10}(d) are the normal brain connections' results. Abnormal brain connections exhibit more instability than normal brain connections in the structural domain. Different with the brain normal connections, the curves of the abnormal connection between brain region Occipital\_Mid\_R and Occipital\_Inf\_R, or brain region Temporal\_Mid\_L and Temporal\_Pole\_Mid\_L show a large degree of amplitude change. And it demonstrates that the abnormal situation got from the experiments are much different with the normal connection. The abnormal brain connections in \{40-56, 85-87, 52-54, 30-72, 41-87\} have the same change trend.

\begin{figure}[h]
	\centering
	\includegraphics[width=0.9\linewidth]{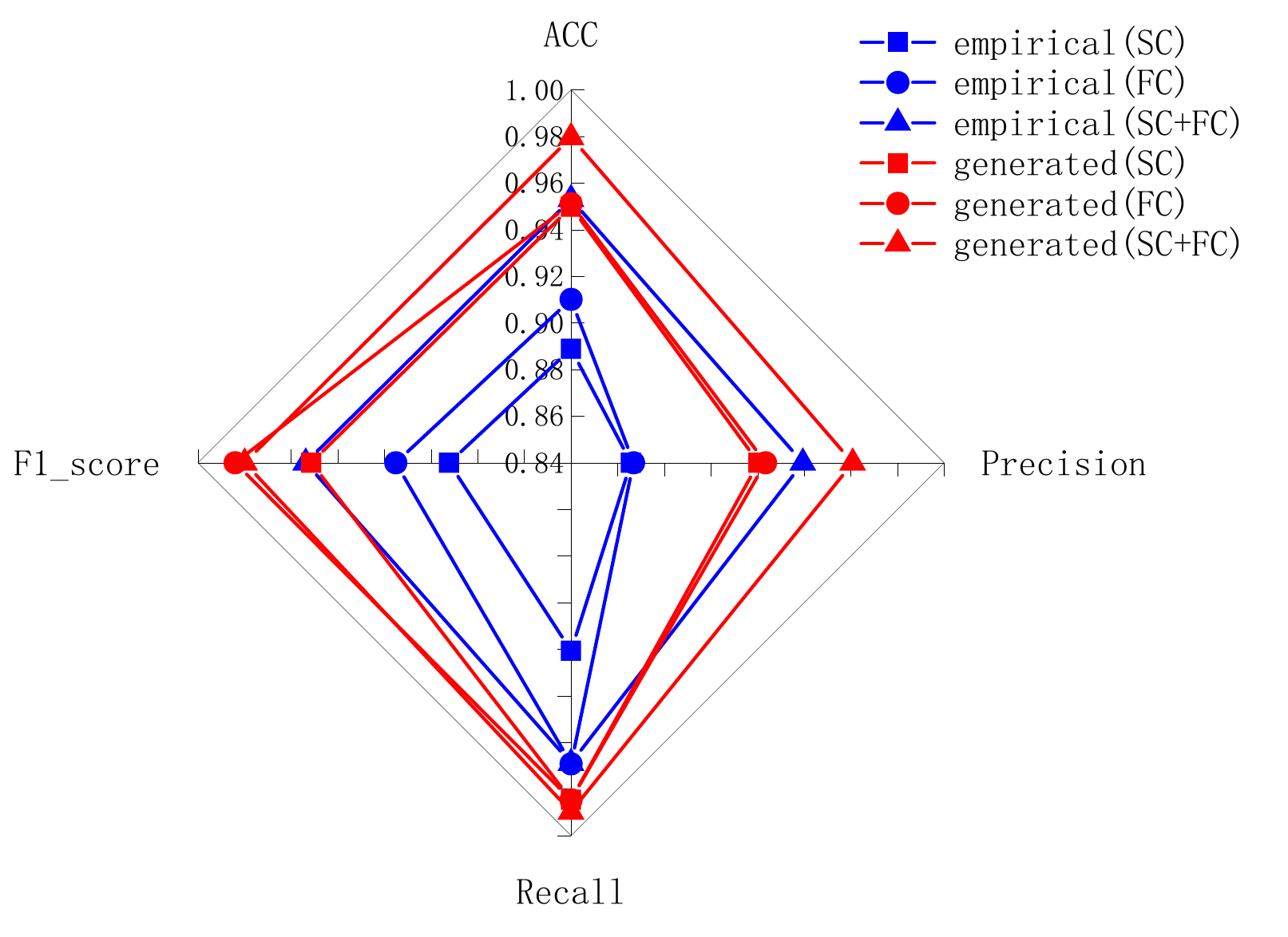}
	\caption{Classification results with the same model but with the different data. The multimodal results are better than the unimodal results, the functional results are better than the structural results and the results for the generated data are better than those for the empirical data. }
	\label{fig:Fig13}
\end{figure}

However, compared with structural data, functional data showed more abnormal results in some brain regions. Among all brain regions, brain regions \{76, 87, 75, 44, 43, 69, 2, 49, 65\} appeared with much frequency.
We used BrainNetViewer\cite{xia2013brainnet} to organize all these abnormal brain connections and brain regions in Fig.\ref{fig:Fig11} and in Fig.\ref{fig:Fig12}.

\subsection{Differences between empirical data and generated data}
The previous subsections have used the generated data for analysis, but we have not compared the difference between the generated data and the empirical data. In this subsection, the classifier is utilized to compare the generated data with the empirical data to demonstrate the effectiveness of generators. After that, the generated data and the empirical data will be analyzed together to get the differences brought by empirical and generated data analysis.

The Fig.\ref{fig:Fig13} shows the classification results obtained with different source data. From the figure, the classification results from the functional data are greater than the structural data, the classification results based on the multiple modalities data are better than the single modal's result and the results from the generated results are better than the empirical data, proving that the generator is indeed optimized for empirical data of Alzheimer's disease. The bidirectional generation attenuates the individual differences between subjects, and because a bidirectional recurrent generative adversarial network is used, the generator can also integrate the data distribution of structure and function to a certain extent, which ultimately leads to the optimization of the data.

\section{Discussion}
\label{Sec:Discussion}
In the above experiments, we compared the differences between structural and functional connections.
When comparing brain structural and functional connection matrices at the individual level, there is not a simple one-to-one correspondence between brain structural and functional connections. There is a high probability that the brain regions with structural connections also have functional connections. However, the brain regions with functional connections do not necessarily have structural connections. It indicates that direct brain structural connectivity does not necessarily have to exist between brain regions that can transmit information and multiple direct brain structural connections can assist in forming indirect brain structure connections to achieve the corresponding functions.

When comparing brain structural and functional connection matrices at the category level, there are large differences in the abnormalities obtained in structure and function. Structural abnormalities were relatively stable. There are some abnormal brain connections existing in all comparative experiments, but there is no function. There are no abnormal brain connections mentioned above, but there are many abnormal brain regions existing in these comparative experiments.

This work hasn't been able to solve all the prevailing issues related to brain structure and connection mapping. Although the correspondence between structure and function can be imitated by generators, the specific correspondence between the structure and function, i.e., the coordination mechanism of the brain, remains ungraspable. Some subjects have large differences compared to the same category of subjects, and how they should go about the diagnosis.

\section{Conclusion}
\label{Sec:Conclusion}

In this paper, we proposed a novel graph convolution method, called InnerGCN, designed to map the structural and functional brain networks of the human brain to one another. The method utilizes two distinct generators and discriminators to learn the structure-to-function and function-to-structure mapping functions, respectively. Our findings indicate that a direct comparison between structural and functional networks does not reveal straightforward similarities, suggesting the absence of a one-to-one correspondence between structure and function. Instead, this highlights the presence of a complex coordination mechanism underlying their interactions. Furthermore, we analyzed the structural and functional connectivity of patients experiencing Alzheimer's Disease (AD) progression. The results show that structural data revealed more stable abnormal brain connections, whereas functional data identified abnormal connections that were less apparent but often associated with specific brain regions. These findings suggest that structural and functional networks provide complementary insights into brain connectivity during disease progression. Future work will explore more advanced coordination models to better capture the complex interactions between structural and functional networks, as well as investigate the generalizability of InnerGCN across other neurodegenerative diseases.

\section*{Acknowledgment}

This work was supported by the National Natural Science Foundations of China under Grant 12326614 and 62172403, the Distinguished Young Scholars Fund of Guangdong under Grant 2021B1515020019.

\bibliographystyle{unsrt}
\bibliography{ref}

\end{document}